\documentclass{article}

\usepackage{PRIMEarxiv}

\usepackage[utf8]{inputenc} 
\usepackage[T1]{fontenc}    
\usepackage{hyperref}       
\usepackage{url}            
\usepackage{booktabs}       
\usepackage{amsfonts}       
\usepackage{nicefrac}       
\usepackage{microtype}      
\usepackage{lipsum}
\usepackage{fancyhdr}       
\usepackage{graphicx}       
\graphicspath{{media/}}     
\usepackage{amsmath}
\usepackage{amssymb}
\usepackage{mathtools}
\usepackage{amsthm}

\title{A Forced-Choice Neural Cognitive Diagnostic Model of Personality Testing
}

\author{
    Xiaoyu Li\textsuperscript{$1$},
    Jin Wu\textsuperscript{$2$,$3$,$4$},
    Shaoyang Guo\textsuperscript{$1$},
    Haoran Shi\textsuperscript{$2$,$3$,$5$},
    \textbf{Chanjin Zheng\textsuperscript{$2$,$3$}\thanks{\ \ Corresponding author.}}
    \\
    \textsuperscript{1}School of Education, Yangzhou University \\    \textsuperscript{2}Lab of Artificial Intelligence for Education, East China Normal University \\
    \textsuperscript{3}Shanghai Institute of Artificial Intelligence for Education, East China Normal University\\
    \textsuperscript{4}School of Computer Science and Technology, East China Normal University \\
    \textsuperscript{5}Department of Educational Psychology, East China Normal University \\
    \texttt{lixiaoyu0228@outlook.com, chjzheng@dep.ecnu.edu.cn} \\
}

\begin{document}
\maketitle

\begin{abstract}

In the smart era, psychometric tests are becoming increasingly important for personnel selection, career development, and mental health assessment. Forced-choice tests are common in personality assessments because they require participants to select from closely related options, lowering the risk of response distortion. This study presents a deep learning-based Forced-Choice Neural Cognitive Diagnostic Model (FCNCD) that overcomes the limitations of traditional models and is applicable to the three most common item block types found in forced-choice tests. To account for the unidimensionality of items in forced-choice tests, we create interpretable participant and item parameters. We model the interactions between participant and item features using multilayer neural networks after mining them using nonlinear mapping. In addition, we use the monotonicity assumption to improve the interpretability of the diagnostic results. The FCNCD's effectiveness is validated by experiments on real-world and simulated datasets that show its accuracy, interpretability, and robustness.
\end{abstract}


\section{Introduction}
In the age of intelligence, psychometric assessments have become increasingly essential. They not only equip individuals with tools for self-awareness but also provide organizations with a scientific foundation for personnel selection\cite{robertson2001personnel,eg2014psychometric}, career development\cite{savickas1996career}, and mental health assessment\cite{mueser2001psychometric}. Psychometric tests are typically divided into cognitive and noncognitive categories. Cognitive tests assess abilities such as logical reasoning and typically use standardized answers, with higher scores indicating greater ability. Intelligent education leverages cognitive testing in various ways, including knowledge tracing \cite{piech2015dkt,abdelrahman2023knowledge}, cognitive diagnosis \cite{wang2022neuralcd,LiuFuzzyCD2018}, and personalized learning \cite{shemshack2020systematic}. Non-cognitive tests, on the other hand, are critical for understanding personality traits, values, and attitudes. They are widely used in clinical diagnosis, career planning, and personnel decisions. Research consistently shows that personality assessments have strong predictive validity for job performance \cite{brown2011opq32r, hurtz2000BF, sitser2013BF}. Likert scales are used in many non-cognitive tests, including the Minnesota Multiphasic Personality Inventory (MMPI) \cite{hathaway1951mmpi} and the Cattell 16 Personality Factor Inventory (16PF) \cite{cattell1956_16PF}. However, in high-stakes situations, participants may be influenced by social desirability, causing them to select responses that do not accurately reflect their true traits, jeopardizing the fairness and discriminability of the outcomes. 

\begin{figure*}[ht]
\vskip 0.2in
\begin{center}
\centerline{\includegraphics[trim=3cm 8cm 5cm 5.6cm, clip,scale=0.7]{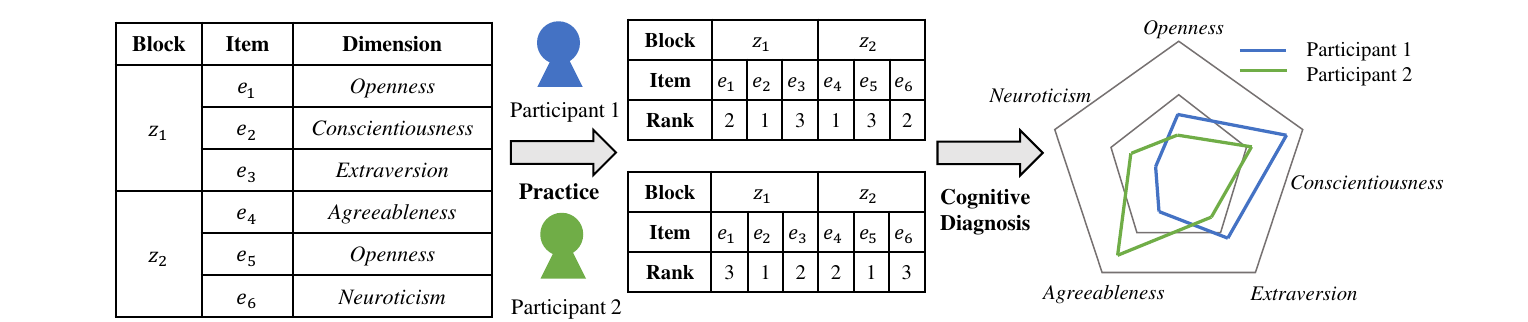}}
\caption{A toy example of forced-choice cognitive diagnosis.}
\label{fc-toy}
\end{center}
\vskip -0.2in
\end{figure*}

To address this issue, researchers have developed forced-choice question formats to effectively mitigate faking and social desirability bias in high-stakes assessments \cite{saville1991,jackson2000,wetzel2021,cao2019}. These tests require participants to choose between two or more options of comparable social desirability, which reduces the possibility of response distortion. PICK, RANK, and MOLE are three common types of forced-choice item blocks. Figure \ref{fc-toy} shows the cognitive diagnosis schematic for the RANK item block type, where participants rate how well the provided descriptions match their own experiences. Based on the participants' responses to the items, an appropriate cognitive diagnostic model was selected to infer their underlying characteristics.

Traditional multidimensional forced-choice (MFC) tests are scored using either conventional scoring techniques or item response theory (IRT). IRT models for forced-choice tests have been developed to determine the relationship between responses and latent traits, generate normative latent trait scores, and allow for inter-individual score comparisons. Common models include Thurstonian Item Response Theory (TIRT) \cite{brown2011tirt} and Multi-Unidimensional Pairwise Preferences (MUPP) \cite{stark2005mupp}. These models typically use Markov Chain Monte Carlo (MCMC) algorithms for parameter estimation. However, this approach can be time-consuming and is limited by assumptions like latent trait linearity and inter-item independence. Furthermore, traditional models frequently struggle with high-dimensional data and complex trait interactions, making it difficult to capture nonlinear relationships and underlying patterns, limiting their practical application.

The rapid advancement of data science and deep learning technologies has made deep learning-based cognitive diagnostic models (CDMs) a popular research topic. These models excel at feature extraction and representation, allowing them to analyze large amounts of sparse data while capturing complex interactions. Several deep learning CDMs have been developed for educational data \cite{WangNCD2020,QiICD2023,liu2024ICDM,GaoDeepCD2022} and Likert type data \cite{li2025pcdf}. However, due to the specific characteristics of forced-choice tests, these existing models cannot be applied directly to forced-choice assessments. As a result, there is an urgent need to create deep learning models specifically designed for forced-choice tests.

To address these issues, this study proposes the Forced-Choice Neural Cognitive Diagnosis (FCNCD) model for evaluating forced-choice personality tests. We start by creating embedding representations for participants and items using their one-hot vectors. These features are then projected into a higher-dimensional space, where feature interactions are computed. The model captures the complex interactions between participants while answering questions using two non-negative fully connected layers that follow the monotonicity assumption. Finally, we use an enhanced BPR loss function to perform pairwise comparisons of items within blocks. Extensive experiments on two real datasets and one simulated dataset show that the FCNCD model is effective at ensuring accuracy and interpretability.

\section{Related Work}
\subsection{Types and Models of Forced-Choice Tests}
Forced-choice tests require individuals to choose two items that best and least represent their characteristics from a list of similar options, or to rank items without the option to positively endorse all choices. These tests consist of multiple blocks with fixed questions and commonly use formats such as PICK (selecting the best match) \cite{oswald2015pick2}, RANK (ranking all items)\cite{brown2011opq32r}, and MOLE (choosing both a MOst and LEast match)\cite{hontangas2015}. The RANK-3 format effectively balances cognitive load and information yield \cite{hontangas2015,joo2018,zheng20242plrank}.

These tests' scoring models are divided into two categories: traditional methods and Item Response Theory (IRT) models. Traditional scoring assigns scores based on response compatibility, which can cause interdependence among scores and complicate statistical analysis \cite{baron1996}. In contrast, IRT models include dominance models like Rasch and two-parameter logistic (2PL) models, which link higher trait levels to more positive responses. Unfolding models, such as the Generalized Grade Unfold Model (GGUM) \cite{roberts2000ggum}, correlate the likelihood of a positive response with item alignment to the assessed trait level. Additional IRT applications include the TIRT \cite{brown2011tirt} model for dominant responses, the MUPP \cite{stark2005mupp} framework for various response modes, models like RIM \cite{wang2017rim} for intra-individual comparisons, and BRB-IRT \cite{lee2020brbirt} for addressing randomized item blocks. These IRT models improve the measurement and analysis capabilities of forced-choice tests, making them more reliable and valid.

\subsection{Ranking algorithm}
Ranking algorithms play an important role in forced-choice tests and are divided into two types: pairwise and listwise methods. Pairwise ranking algorithms optimize models by comparing user preferences in pairs. One notable example is Bayesian Personalized Ranking (BPR) \cite{rendle2012bpr}, which improves ranking accuracy by maximizing score differences between high and low preference items, making it appropriate for implicit feedback. Other classic methods include RankNet \cite{burges2005ranknet}, which minimizes the gap between predicted and actual rankings, and LambdaRank \cite{burges2006lambdarank}, which directly optimizes information retrieval metrics. In educational contexts, researchers frequently use non-interactive exercise data to create pairs of positive and negative samples, optimizing them with BPR loss, as seen in the EIRS model \cite{yaoEIRS2023} and the CMES model \cite{maCMESlongtail2024}. In contrast, listwise ranking methods optimize the entire list's sorting structure. ListNet \cite{cao2007listnet} optimizes using KL divergence, while ListMLE \cite{xia2008listmle} uses maximum likelihood estimation, AdaRank \cite{xu2007adarank} iteratively tunes parameters, and LambdaMART \cite{burges2010lambdamart} combines gradient boosting trees with LambdaRank. These methods provide a more comprehensive view of global rankings.

\section{Research Background}
Here is a formal definition of forced-choice test. Suppose there are $N$ participants, $M$ Items, $K$ 
dimensions and $L$ item block in a test, denoted $S=\{s_1,s_2,...,s_N\}$, $E=\{e_1,e_2,...,e_M\}$, $C=\{c_1,c_2,...,c_K\}$, and $Z=\{z_1,z_2,...,z_L\}$, where each item block $z_l$ contains $t$ items, i.e., $z_l \subseteq E$. During the answering process, each participant $s_n$ ranks the items in item block $z_l$ based on how well the descriptions fit them. The participant's response log is denoted as $R=(s_{n},z_{l},r_{n,l})$, where $r_{n,l}$ represents the ranking result for item block $z_l$. 

The format of $r_{n,l}$ varies by item block type:
\begin{itemize}
    \item \textbf{PICK:} Participants select the most conforming item from item block $z_l$. This item is assigned a value of $t$, while all other items receive a value of 1. 
    
    \item \textbf{RANK:} Participants rank all items in item block $z_l$ are ranked by participants in full order.  in order of compatibility. The most compatible item scores $t$, the second most compatible scores $t-1$, and the least compatible item scores 1.
    
    \item \textbf{MOLE:} Participants choose one most conforming and one least conforming item from item block $z_l$. The most conforming item is assigned a value of 3, the least conforming a value of 1, and all other items receive a value of 2.
\end{itemize}

In addition, the test has a Q-matrix $Q=\{Q_{mk}\}_{M \times K}$, where $Q_{mk}=1$ if item $e_m$ relates to dimension $c_k$ and $Q_{mk}=0$ otherwise.

\textbf{Problem Definition:} Given the participants' response logs $R$ and the Q-matrix $Q$, our goal is to predict the ordering of items within an item block as closely as possible to the participants' true ranking results.

\section{Forced-Choice Neural Cognitive Diagnostic Model}
\begin{figure*}[ht]
\vskip 0.2in
\begin{center}
\centerline{\includegraphics[trim=2.9cm 5.5cm 3.5cm 4.2cm, clip,scale=0.7]{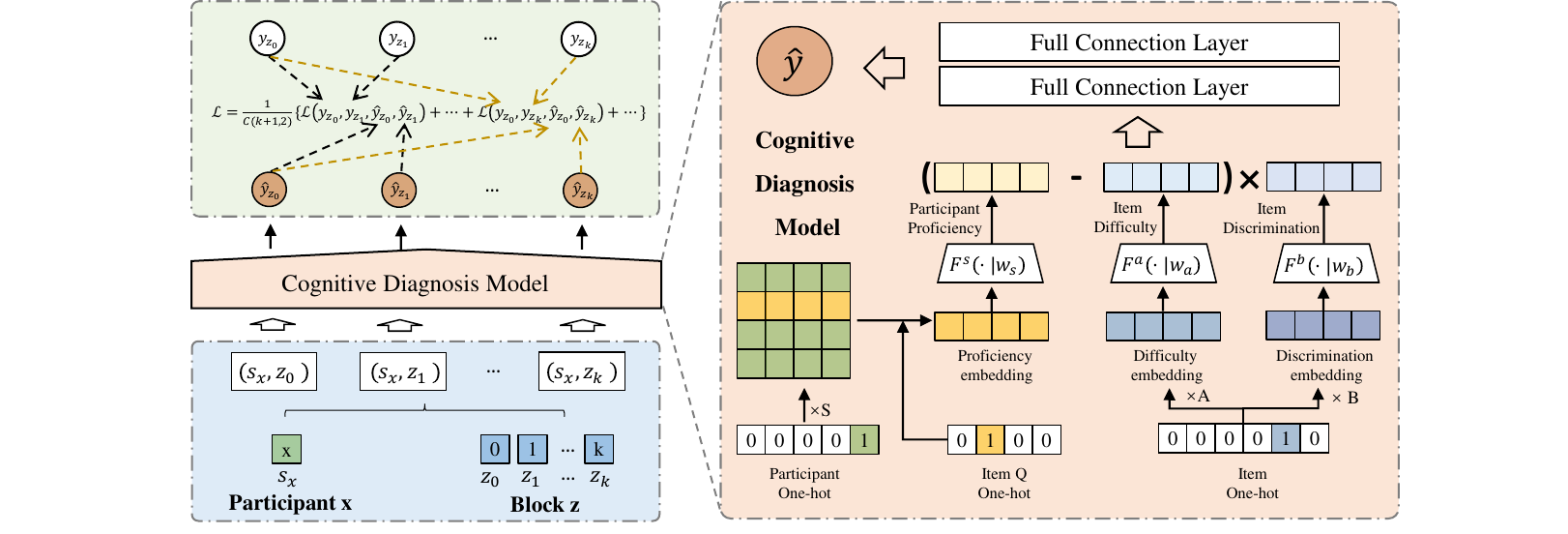}}
\caption{The framework of the proposed FCNCD model.}
\label{FCNCD framework}
\end{center}
\vskip -0.2in
\end{figure*}

To enhance the assessment of forced-choice tests, this paper introduces the Forced-Choice Neural Cognitive Diagnosis (FCNCD) model. Figure \ref{FCNCD framework} illustrates the framework of the FCNCD model.

\subsection{Embedded Representation of Participant and Item Factors}
In cognitive diagnosis, it is common to characterize both participant and item factors.

\textbf{Participant Factors.} In FCNCD, each participant is represented by a proficiency matrix.  The participant factor $\boldsymbol{s}$ is derived by multiplying the participant's one-hot representation vector $\boldsymbol{x}_s$ with a trainable matrix $\boldsymbol{W}_s$:
\begin{equation}
    \boldsymbol{s} = \boldsymbol{x}_s \times \boldsymbol{W}_s,
\end{equation} in which $\boldsymbol{s} \in \mathbb{R}^{1 \times K \times d}$, $\boldsymbol{x}_s \in \{0,1\}^{1 \times N}$, $\boldsymbol{W}_s \in \mathbb{R}^{N \times K \times d}$.
The participant ability dimension feature presentation corresponding to the item is denoted as:
\begin{equation}
    \boldsymbol{h}_1^{\text{prof}} = \boldsymbol{s} \odot \text{unsqueeze}(\boldsymbol{x}_q) ,
\end{equation}
where $\boldsymbol{h}_1^{\text{prof}} \in \mathbb{R}^{1\times d}$ denotes the participant ability dimension feature presentation corresponding to the item. Here, an unsqueeze operation is applied to the second dimension of $\boldsymbol{x}_q$ to ensure compatibility for element-wise multiplication, and $\text{unsqueeze}(\boldsymbol{x}_q) \in \{0,1\}^{1 \times K \times 1}$.

\textbf{Item Factors.} As for each item, item difficulty and discrimination are defined as follows:
\begin{equation}
    \boldsymbol{h}_1^{\text{diff}} = \boldsymbol{x}_e \times \boldsymbol{W}^{\text{diff}}, 
\end{equation}
\begin{equation}
    \boldsymbol{h}_1^{\text{disc}} = \boldsymbol{x}_e \times \boldsymbol{W}^{\text{disc}},
\end{equation}
where $\boldsymbol{h}_1^{\text{diff}}, \boldsymbol{h}_1^{\text{disc}} \in \mathbb{R}^{1\times d}$ are the embedded representations of item difficulty and discrimination, respectively. Here, $\boldsymbol{x}_e \in \{0,1\}^{1 \times M}$ is the one-hot representation of the item, while $\boldsymbol{W}^{\text{diff}}, \boldsymbol{W}^{\text{disc}} \in \mathbb{R}^{M \times d}$ are the learnable matrices.

\subsection{Non-linearity Mapping Layer}
After obtaining the embedded representations, deep feature extraction is performed on the participant's ability, item difficulty, and discrimination using a nonlinear activation function. This procedure introduces nonlinear representations of the features. Specifically, the transformations are defined as follows:
\begin{equation}
    \boldsymbol{h}_2^{\text{prof}} = \phi (\boldsymbol{W}_1 \times \boldsymbol{h}_1^{\text{prof}} + \boldsymbol{b}_1),
\end{equation}
\begin{equation}
    \boldsymbol{h}_2^{\text{diff}} = \phi (\boldsymbol{W}_2 \times \boldsymbol{h}_1^{\text{diff}} + \boldsymbol{b}_2),
\end{equation}
\begin{equation}
    \boldsymbol{h}_2^{\text{disc}} = \phi (\boldsymbol{W}_3 \times \boldsymbol{h}_1^{\text{disc}} + \boldsymbol{b}_3),
\end{equation} where $\boldsymbol{W}_1$, $\boldsymbol{W}_2$, $\boldsymbol{W}_3$ are the learnable matrices, $\boldsymbol{b}_1$, $\boldsymbol{b}_2$, $\boldsymbol{b}_3$ are the corresponding bias terms, and $\phi$ denotes the activation function, which is the Sigmoid function in this paper.

\subsection{Interaction Function}

The interaction function is defined as:
\begin{equation}
    \boldsymbol{x} = \boldsymbol{h}_2^{\text{disc}} \times (\boldsymbol{h}_2^{\text{prof}} - \boldsymbol{h}_2^{\text{diff}}),
\end{equation}

This is followed by a fully connected layer and an output layer:
\begin{equation} \label{mono1}
    \boldsymbol{f}_1 = \phi (\boldsymbol{W}_4 \times \boldsymbol{x} + \boldsymbol{b}_4),
\end{equation}
\begin{equation} \label{mono2}
    {y} = \phi (\boldsymbol{W}_5 \times \boldsymbol{f}_1 + b_5).
\end{equation}
To satisfy the monotonicity assumption, we we constrain each element of $\boldsymbol{W}_4$ and $\boldsymbol{W}_5$ to be nonnegative. Here, $\boldsymbol{b}_4$ and $\boldsymbol{b}_5$ are the corresponding bias terms, $\phi$ is the activation function, and ${y}$ represents the final output of the model.

\subsection{Prediction of Scores for Items in Item Block}
In a forced-choice test, the model's inputs are participant $s_n$ and item block $z_l$. It applies to each item within the block $z_l$ and predicts scores for all items:
\begin{equation}
    {y}_{n,l} = \{{y}_{n,l_1},{y}_{n,l_2},...,{y}_{n,l_t}\}.
\end{equation}

The ranking score of the items in the item block is given by:
\begin{equation}
    p_{n,l} = rank(y_{n,l}),
\end{equation}
where $rank(y_{n,l})$ represents the ranked position of each probability value in the set $y_{n,l}$. The RANK item block type ranks based on probability values, with the smallest value assigned a rank of 1 and the largest a rank of $t$. In contrast, the ranking system for the MOLE item block type is slightly different: the smallest probability value is ranked first, the largest is ranked third, and the remaining values are ranked second.

\subsection{Model Optimization}
Our method classifies items with higher ranking positions as positive samples and those with lower rankings as negative samples. In CDM, this means that higher-ranked items should have higher predicted scores than lower-ranked items. We optimize ranking using the BPR loss function \cite{rendle2012bpr} from recommender systems, which maximizes the score difference between positive and negative samples. To better suit the ranking task of forced-choice tests, we improve the BPR loss by using the difference in ranking scores as a weighting variable. The BPR loss for each pair of items in the item block is as follows:
\begin{equation}
    \mathit{L}(y_i, y_j, r_i, r_j) = -\ln \big( \sigma \big( \frac{\lambda}{r_i - r_j} \cdot (y_i - y_j) \big) \big)
\end{equation}
where $y_i$ and $y_j$ are the predicted scores for items $e_i$ and $e_j$, respectively, $r_i$ and $r_j$ are their ranked positions, $\lambda$ is a weighting coefficient, and $\sigma(\cdot)$ is the Sigmoid activation function. This loss function aims to ensure that top-ranked items score higher than those ranked lower.

For the RANK item block type, the total loss is given by:
\begin{equation}
    \text{loss}_{\text{RANK}} = \frac{1}{\binom{t}{2}} \sum_{1 \leq i < j \leq t} L(y_i, y_j, r_i, r_j),
\end{equation}
where $\binom{t}{2}$ represents the number of combinations (i.e., pairs of items) selected from the $t$ items in the block, and $L(y_i, y_j, r_i, r_j)$ is the BPR loss for items $e_i$ and $e_j$. This formula calculates the average loss across all item pairs.

For the MOLE item block type, the total loss is expressed as:
\begin{equation}
    \text{loss}_{\text{MOLE}} = \frac{1}{\binom{t}{2} - \binom{t-2}{2}} \sum_{1 \leq i < j \leq t} \mathbb{I}_{\text{MOLE}}(r_i - r_j) \, L(y_i, y_j, r_i, r_j),
\end{equation}
where $\mathbb{I}_\text{MOLE}(\cdot)$ is the indicator function that equals 0 when $r_i = r_j$ and 1 otherwise. The $\text{loss}_{MOLE}$ is calculated by considering only pairs of items with different ranks, effectively excluding those with equal ranks.

\section{Experiments}
We conduct comprehensive experiments to explore the following research questions:
\begin{itemize}
    \item \textbf{RQ1:} How sensitive is FCNCD to its model parameters?
    \item \textbf{RQ2:} Does the FCNCD model outperform the baseline models?
    \item \textbf{RQ3:} How effective are the key components of the FCNCD model?
    \item \textbf{RQ4:} How interpretable are participants' diagnostic states in the FCNCD model?
    \item \textbf{RQ5:} What do the visualizations show about the interactions between participants, items, and dimensions?
\end{itemize}

\subsection{Datasets}
The experiments are carried out with two real-world datasets and one simulated dataset: MAP, BFI, and sim-mole. 


\begin{itemize}

    \item \textbf{MAP} is a business personality test developed by a company, consisting of 264 items that assess 24 dimensions across three aspects: Mental (M), Attitudes and Motivation (A), and People Skills (P). The test features 88 item blocks, with each block containing three items from different dimensions and a total of 11 items per dimension. Complete response data were collected from 1,433 participants, organized in an item block format using RANK.

    \item \textbf{BFI} is a forced-choice personality assessment based on the Big Five Inventory 2 \cite{soto2017bfi}. It includes 60 items across five original dimensions. In this study, positively and negatively scored items are treated as separate dimensions, resulting in a total of 10 dimensions and 20 item blocks, each containing three items from different dimensions. The dataset includes responses from 372 participants, also formatted as RANK item blocks.

    \item \textbf{sim-mole} dataset comprises 480 items across 24 dimensions and 120 item blocks, with each block containing four questions from different dimensions. Responses were generated for 1,000 participants under specific conditions: the discrimination parameter was randomly sampled from a uniform distribution $U(0.75,1.5^2)$, and the difficulty parameter from a normal distribution $N(0,0.5^2)$. For each participant, a 24-dimensional vector of latent traits was generated from a multivariate normal distribution with a mean of 0 and a fixed covariance of 0.5 between dimensions. Participant responses were simulated using the MOLE response model, employing the Luce decision-making approach to calculate the probabilities of each response mode. Based on these probabilities, response matrices were generated using a multinomial distribution. 

\end{itemize}

Table \ref{FCNCD dataset} provides detailed statistics for these three datasets. 

\begin{table}[h]
\caption{Dataset summary.}
\label{FCNCD dataset}
\vskip 0.15in
\begin{center}
\begin{tabular}{llll}
\toprule
   &MAP&	BFI  &sim-mole\\
\midrule
\# Dimension              &	24      &	10  &24   \\
\# Participants            &	1433    &	372  &   1000   \\
\# Items          &	264     &	60   &    480   \\
\# Item blocks&	88      &	20   &120      \\
Items per item block      & 3        &    3   &4       \\
Item block type          &	RANK    &   RANK  &MOLE \\
\bottomrule
\end{tabular}
\end{center}
\vskip -0.1in
\end{table}

\subsection{Baselines}
This study employs several baseline methods, including MUPP-2PL, BRB-IRT, MF, RankNet, NCDM-R, KaNCD-R, CDND-R, and Random.

\begin{itemize}
    \item \textbf{MUPP-2PL} ~\cite{morillo2016mupp2pl}  is developed from the dominant response model and integrated with the MUPP framework \cite{stark2005mupp} for effective selection of question type data. Its formula is given by:
    \begin{equation}
            P(i>j|\theta_{q_i},\theta_{q_j}) = \phi(a_i \theta_{q_i} -a_j\theta_{q_j}+a_ib_i-a_jb_j),
    \end{equation} where $\phi$ denotes the sigmoid function, also known as the logistic Stick function in educational psychology. Here, $a_i$ and $a_j$are discrimination parameters for items, $\theta_{q_i}$ and $\theta_{q_j}$represent the latent traits for items $i$ and $j$, respectively, and $a_ib_i-a_jb_j$ serves as the intercept parameter derived from combining the $a$ and $b$ parameters in the 2PLM.
    
    MUPP-2PL was applied to the PICK-2 forced-choice test. To adapt the tasks of this study, data from RANK and MOLE question types were converted to the PICK-2 format. The same data processing method was applied to the other benchmark methods described below.
    
    \item \textbf{BRB-IRT} ~\cite{lee2020brbirt} is a Bayesian item block model \cite{bradlow1999bayesian} selected as the base model for forcing question selection. It incorporates the randomized question block effect parameter $\gamma_l$ into the item response function of MUPP-2PL, considering the interdependence of topics within the block during parameter estimation. Its formula is:

    \begin{equation}
            P_{n,l}(i>j|\theta_{q_i},\theta_{q_j})  =
            \phi(a_i\theta_{q_i}-a_j\theta_{q_j} -(a_ib_i-a_jb_j)-\gamma_{n,l}),
    \end{equation}

    where $\gamma_{n,l}$ represents the random block effect for participant $n$ on item block $l$, reflecting the influence of the dimensions measured by that block on the participant's response.

    \item \textbf{MF} ~\cite{WangNCD2020}  is a matrix decomposition model used as a baseline in various cognitive diagnosis studies in education. It aligns participants and items with users and items in matrix decomposition, effectively predicting participant performance. However, it lacks explanatory power for cognitive diagnosis, as there is no clear mapping between elements in the trait vector and specific dimensions. Its formulation is given by:
    \begin{equation} 
    \label{mf}
        y = MLP( \boldsymbol h_e \circ \boldsymbol h_s),
    \end{equation} where $\boldsymbol h_e$ is the latent trait vector for the topic, $\boldsymbol h_s$ is the latent trait vector for the subject, $\circ$ represents element-wise multiplication, and $MLP$ consists of two fully connected layers and one output layer.

    \item \textbf{RankNet} \cite{burges2005ranknet} employs a neural network to predict relative rankings between pairs of items, aiming to minimize the difference between predicted and true rankings. It computes relative preferences by comparing pairs of items (positive and negative samples) and is trained using a cross-entropy loss function. The network structure of RankNet is similar to that of MF, as shown in Eq. \ref{mf}, but it uses the ReLU activation function in the fully connected layer, while the prediction layer remains Sigmoid. The probability of item $i$ being ranked higher than item $j$ is expressed as:
    \begin{equation}
    \label{rankmethod}
        P(i>j) = \phi(y_i-y_j),  
    \end{equation}

    \item \textbf{NCDM-R} \cite{WangNCD2020} is a widely used deep learning-based cognitive diagnostic model that employs a multilayer perceptron to capture complex, higher-order interactions between subjects and topics. Its formulation is:
    \begin{equation} 
        y=MLP(\mathbf{Q}_e\circ(\mathbf{h}_s-\mathbf{h}^{diff})\times h^{disc}),
    \end{equation} where $\mathbf{h}^{diff}$ and $h^{disc}$ represent the difficulty of knowledge concepts and topic differentiation, respectively, and $\mathbf{h}_s$ denotes the subject's proficiency. The $MLP$ consists of two fully connected layers and one output layer. The original NCDM only provides predicted scores for questions without ranking. To adapt it for forced-choice item types, we apply the ranking probability formula from RankNet, resulting in NCDM-R. 
    \item \textbf{KaNCD-R} \cite{wang2022neuralcd} builds on NCDM and incorporates the implicit relationships between knowledge concepts.
    \begin{equation}
        y=MLP(\mathbf{Q}_e\circ(\mathbf{h}_s \cdot \mathbf{l}-\mathbf{h}^{diff} \cdot \mathbf{l})\times h^{disc}),
    \end{equation} where $\mathbf{l}$ represents the vector of knowledge concepts associated with item $e$. For the modification of KaNCD, we adopt the same approach as in NCDM.
    \item \textbf{CDND-R} \cite{zhang2025cognitive} uses an attention-like mechanism to capture the nonlinear dependencies between participants and items, while employing a discriminative interaction approach to exploit second-order interactions, thereby improving discriminative capability across different participant-item combinations. The CDND-R modifications take the same approach as those for NCDM.
    \item \textbf{Random} method predicts the participant's probability of success on a topic using a uniform distribution $Uniform(0,1)$. This generates a ranked score based on the magnitude of the predicted probabilities for each topic pair. In the interpretability analysis, the ability values of subjects across different dimensions are also randomly generated from the uniform distribution $Uniform(0,1)$.
    \end{itemize}

Among the baseline models, MF was optimized using the BPR loss function, while the other models were optimized with the cross-entropy loss function.

\subsection{Experimental Details}
For parameter initialization, we used the Xavier method \cite{glorot2010xavier}, and the AdamW optimizer \cite{loshchilov2017adamw}. The embedding dimension $d$ was set to 64, with the nonlinear mapping layer sized at 256 and the fully connected layer at 128. Experiments were conducted on a Linux server equipped with a 2.50 GHz Xeon Platinum 8255C CPU, a Tesla T4 GPU, and 32 GB of RAM. Each experiment was repeated 10 times, and the final results represent the average of these repetitions.

\subsection{Evaluation Metrics}
The forced-choice personality test asks participants to rank all items in an item block based on their perceptions. This study compared model performance to the accuracy of item rankings using two metrics: pairwise rank accuracy (PRA) and listwise rank accuracy (LRA).

\textbf{PRA} measures the consistency between the model's predicted rankings and the actual rankings across all possible pairs of items. The definition is as follows:
\begin{equation}
    \text{PRA} = \frac{1}{L} \sum_{l=1}^{L} \frac{1}{\binom{t}{2}} \sum_{1 \leq i < j \leq t} \mathbb{I}_{\text{PRA}} \left( (y_i > y_j) \land (r_i > r_j) \right),
\end{equation}

where $L$ is the total number of item blocks, $t$ is the number of items in each block, $e_i$ and $e_j$ are the indices of the item pairs, ${\binom{t}{2}}$ represents the total number of pairwise combinations, and $\mathbb{I}_{\text{PRA}}(\cdot)$ is the indicator function. A pair is considered correct when the predicted score ordering aligns with the actual ordering.

\textbf{LRA} evaluates the accuracy of the model's predicted ranking of the entire item block, using the following formula:
\begin{equation}
    \text{LRA} = \frac{1}{L} \sum_{l=1}^{L} \mathbb{I}_{\text{LRA}} \left( p_l = r_l \right),
\end{equation}
where the predicted orderings $p_l$ and the actual ranking $r_l$ must be identical for a correct classification.

\subsection{Hyperparameter Analysis (RQ1)}
The hyperparameters in FCNCD include $\lambda$, batch size, and training set ratio. This section examines how these hyperparameters affect FCNCD's performance and assesses its robustness. 

\subsubsection{The effect of $\lambda$}
\begin{figure}[ht]
\vskip 0.2in
\begin{center}
\centerline{\includegraphics[trim=0cm 0cm 0cm 0cm, clip,scale=0.6]{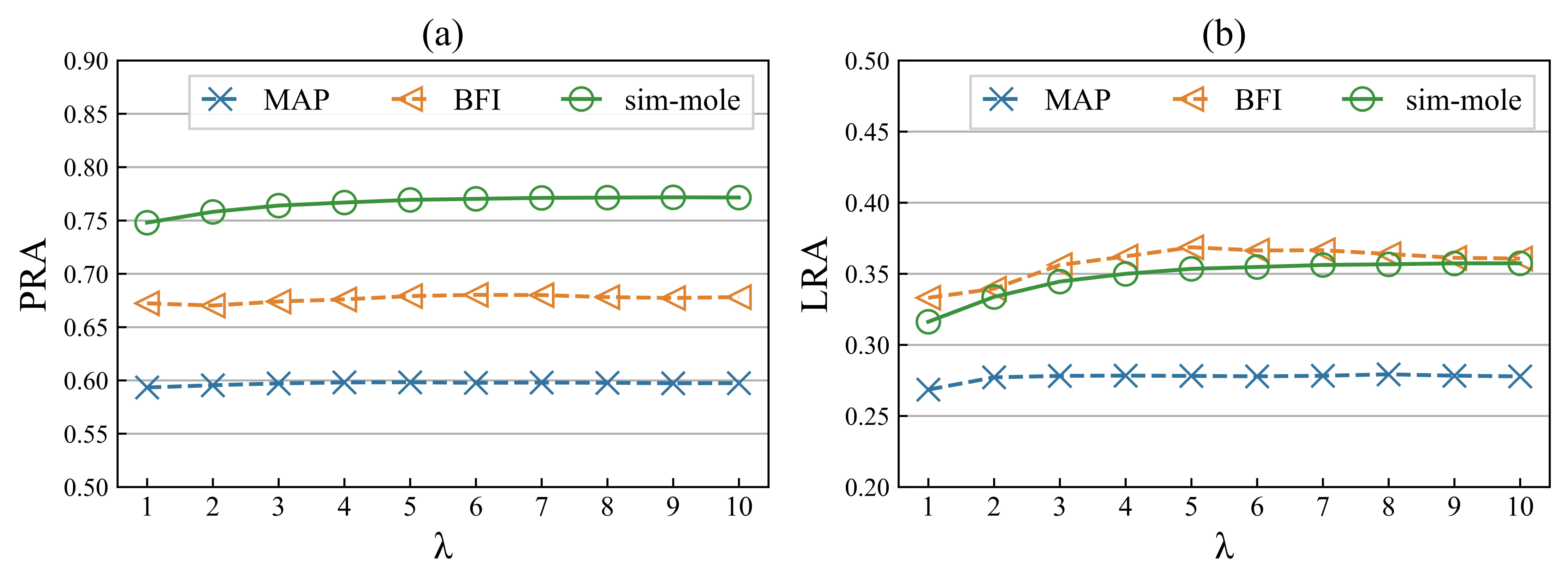}}
\caption{The impact of $\lambda$.}
\label{lambda-sensitivity}
\end{center}
\vskip -0.2in
\end{figure}

The analysis of the hyperparameter $\lambda$ is presented in Figure\ref{lambda-sensitivity}. For the MAP dataset, both the PRA and LRA show minimal variation with changes in $\lambda$, indicating low sensitivity, with optimal performance achieved at $\lambda = 8$. In the BFI dataset, PRA and LRA initially increase before leveling off, peaking around $\lambda = 5$. This suggests that a moderate $\lambda$ value enhances model optimization. Conversely, in the sim-mole dataset, PRA and LRA consistently increase with $\lambda$,  particularly between $\lambda = 1$ and $\lambda = 5$, where performance improvements are most pronounced. However, for $\lambda > 5$, the growth in PRA and LRA diminishes and stabilizes, indicating that larger $\lambda$ values yield limited performance gains.

Overall, the response to $\lambda$ varies across datasets: a moderate $\lambda$ optimizes performance in real-world datasets, while a larger $\lambda$ significantly improves the simulated dataset's performance, albeit with diminishing returns. This highlights the need for tailored $\lambda$ values to optimize model performance across different datasets.

\subsubsection{The effect of batch size}
\begin{figure}[ht]
\vskip 0.2in
\begin{center}
\centerline{\includegraphics[trim=0cm 0cm 0cm 0cm, clip,scale=0.6]{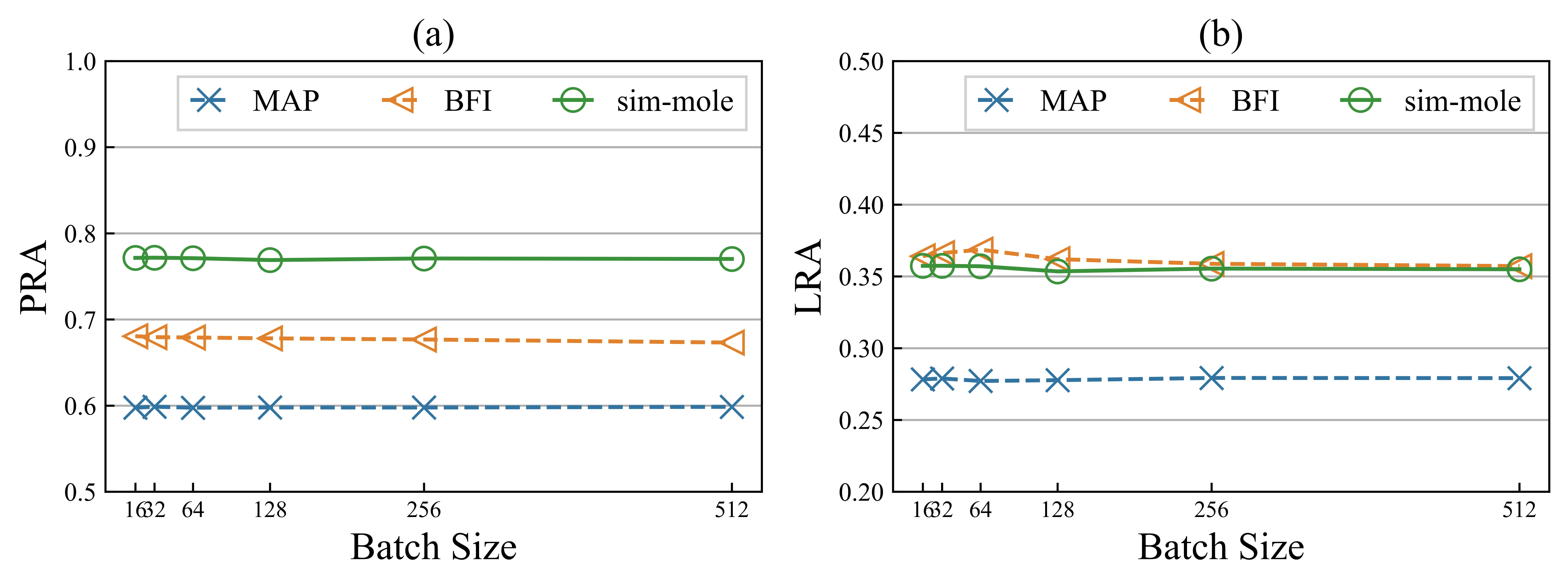}}
\caption{The impact of batch size.}
\label{batchsize-sensitivity}
\end{center}
\vskip -0.2in
\end{figure}
The analysis of batch size is illustrated in Figure \ref{batchsize-sensitivity}. For the MAP dataset, both PRA and LRA remain stable across varying batch sizes, indicating that the dataset can sustain consistent model performance. This stability facilitates practical applications and allows for training with larger batches, thereby enhancing training efficiency. In contrast, the BFI and sim-mole datasets perform optimally with smaller batches, as larger batches result in performance degradation.

\subsubsection{The effect of train set ratio}
\begin{figure}[ht]
\vskip 0.2in
\begin{center}
\centerline{\includegraphics[trim=0cm 0cm 0cm 0cm, clip,scale=0.6]{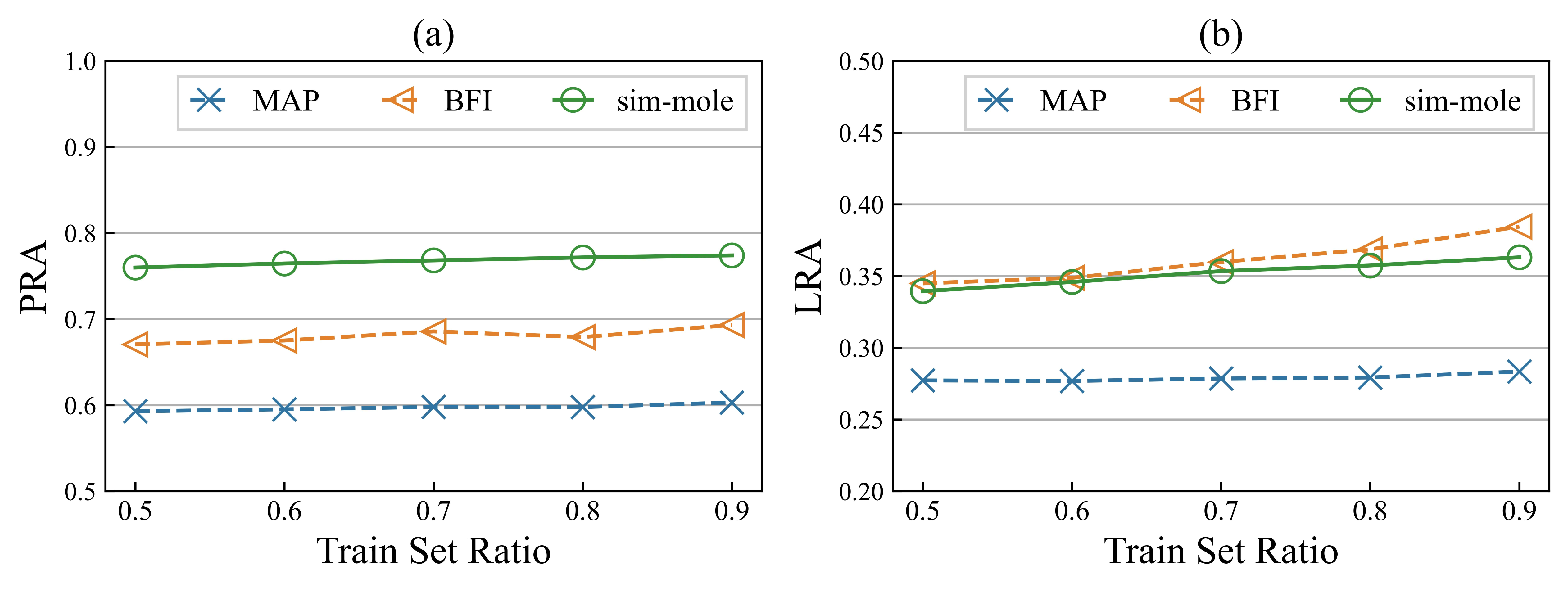}}
\caption{The impact of train set ratio.}
\label{test-ratio-sensitivity}
\end{center}
\vskip -0.2in
\end{figure}

The analysis of the training set ratio is presented in Figure \ref{test-ratio-sensitivity}. The MAP dataset demonstrates strong stability, remaining largely unaffected by variations in the training set ratio. In contrast, both the BFI and sim-mole datasets show a gradual performance increase with higher training set ratios.

The final hyperparameters for the MAP, BFI, and sim-mole datasets are as follows: $\lambda$ values of 8, 5, and 10; batch sizes of 256, 64, and 32; and learning rates of 1e-2, 5e-3, and 5e-4 respectively. Participant response data was randomly divided into training and test sets in an 8:2 ratio by item block. An early-stopping mechanism was implemented: training stops and the best model is saved if evaluation metrics do not improve for 5 consecutive iterations.

\subsection{Performance Prediction Task for Participants (RQ2)}
This paper evaluates cognitive diagnostic results of various models through the participant performance prediction task, summarized in Table \ref{Performance FCNCD}. The table's bolded entries represent the best results for each metric, while underlined entries represent suboptimal results. The analysis leads to the following conclusions: \textbf{(1)} The FCNCD model performs well across all metrics. With the exception of a suboptimal PRA index in the BFI dataset, its PRA and LRA scores outperform other benchmark models in the remaining datasets, demonstrating its efficacy and dependability in predicting participant performance on forced-choice tests. \textbf{(2)} NCDM-R and KaNCD-R perform well in comparison to the benchmark models. NCDM-R ranks second to FCNCD on the MAP and sim-mole datasets, while KaNCD-R ranks second in LRA metrics for the BFI dataset, demonstrating their competitiveness in specific scenarios.
\textbf{(3)} Additional analysis reveals that neural network-based models outperform statistical models on the MAP and BFI datasets with RANK-3 item block types. In contrast, the sim-mole dataset, which uses MOLE-4 item block types, favors statistical models. This implies that model performance is influenced by item block types and data characteristics, with different methods demonstrating strengths and weaknesses in different contexts.
In conclusion, the FCNCD model outperforms existing benchmark models in terms of overall performance while also demonstrating strong predictive capability for forced-choice tests.

\begin{table}[h]
\caption{Experimental results of participant performance prediction.}
\label{Performance FCNCD}
\vskip 0.15in
\begin{center}
\begin{tabular}{lllllll}
\toprule
& \multicolumn{2}{c}{MAP} &\multicolumn{2}{c}{BFI} & \multicolumn{2}{c}{sim-mole} \\
 \midrule
&PRA & LRA & PRA & LRA & PRA & LRA \\
 \midrule
Random&0.500&	0.166&	0.498&	0.164&	0.500&	0.083\\
MUPP-2PL&	0.560&	0.249&	0.605&	0.271&	0.748&	0.319\\
BRB-IRT&0.567&	0.253&	0.607&	0.279&	0.748&	0.316\\
MF& 0.573&	0.264&	0.649&	0.311&	0.689&	0.243\\
RankNet&	0.577&	0.265&	0.651&	0.317&	0.686&	0.238\\
NCDM-R&	{0.596}&	0.270&	{0.681}&	0.349&	{0.749}&{0.319}\\
KaNCD-R&	0.595&	0.268&	\textbf{0.681}&	{0.352}&	0.739&	0.308\\
CDND-R &    0.583&   {0.274}&   0.669&   0.339&   0.699&   0.250\\
FCNCD&\textbf{0.598}&\textbf{0.279}&0.679&\textbf{0.369}& \textbf{0.772}&	\textbf{0.357}\\
\bottomrule
\end{tabular}
\end{center}
\vskip -0.1in
\end{table}

\subsection{Ablation experiments (RQ3)}
To assess the contribution of each model component to overall performance, we performed ablation experiments, creating three variants:
\begin{itemize}
    \item \textbf{FCNCD\_EB}: This variant removes the non-linearity mapping layer, directly constructing the interaction function using $\boldsymbol{h}_2^{\text{prof}}$, $\boldsymbol{h}_2^{\text{diff}}$, and $\boldsymbol{h}_2^{\text{disc}}$ instead of $\boldsymbol{h}_1^{\text{prof}}$, $\boldsymbol{h}_1^{\text{diff}}$, and $\boldsymbol{h}_1^{\text{disc}}$ to assess the impact of the non-linearity mapping layer.
    \item \textbf{FCNCD\_BPR}: This variant uses the original BPR loss function but excludes the difference in ranked scores as a weighting term. The formula for calculating the original BPR loss is:
    \begin{equation}
        \mathit{L}(y_i, y_j, r_i, r_j) = -\ln \sigma \left( \text{sgn}(r_i - r_j) \cdot (y_i - y_j) \right)
    \end{equation} where $\text{sgn}$ is the \textbf{sign} function, which returns the sign of the input. 
    
    \item \textbf{FCNCD\_List}: This variant replaces BPR loss with list loss, which calculates the logarithmic difference between the predicted scores of each item in the block and their subsequent output scores. The predicted scores are logarithmized and compared with the exponential values of the remaining items, taking the negative average. The specific formula for list loss is:
    \begin{equation} 
        \mathit{L} = -\frac{1}{T} \sum_{i=1}^{T} \big( y_{n,i} - \log \big( \sum_{j=i}^{T} e^{y_{n,j}} \big) \big),
    \end{equation}
    where $T$ is the number of items in the block and $y_{n,i}$ is the predicted score for the $i$-th item. This loss function promotes global consistency in sequence predictions while enhancing the model's focus on overall ranking.
    
    \item \textbf{FCNCD\_MO}: To investigate the impact of the monotonicity assumption on model performance, FCNCD\_MO removes the non-negative constraints on $\boldsymbol{W}_4$ and $\boldsymbol{W}_5$ in Equations \ref{mono1} and \ref{mono2}.
\end{itemize}

Table \ref{ablation} presents the ablation experiment results. The key findings include: \textbf{(1)} The complete FCNCD model outperforms all variants, implying that the components work together to improve predictive ability. Furthermore, FCNCD\_MO produces results similar to FCNCD, indicating that the monotonicity assumption has little effect on predictive performance. As a result, FCNCD\_MO will be included in future interpretability analyses. \textbf{(2)} Removing the non-linearity mapping layer in FCNCD\_EB reduces performance across all metrics, indicating its effectiveness in identifying participant and item factors. \textbf{(3)} The performance of the improved BPR loss and pairwise comparison modules varies according to item block type. In the RANK item block type, FCNCD\_List shows significant performance degradation in the PRA metric, while the decline in the LRA metric is less severe. FCNCD\_BPR primarily improves LRA metrics and has little impact on PRA metrics. FCNCD\_List performs significantly worse than FCNCD\_BPR for the MOLE item block type, likely due to incomplete sorting. This complicates the global ranking computation and hinders model performance. The pairwise comparison method is more effective for dealing with incomplete ranking tasks.

\begin{table}[h]
\caption{Ablation experiments.}
\label{ablation}
\vskip 0.15in
\begin{center}
\begin{tabular}{ccccccc}
\toprule
& \multicolumn{2}{c}{MAP} &\multicolumn{2}{c}{BFI} & \multicolumn{2}{c}{sim-mole} \\
\midrule
&PRA & LRA & PRA & LRA & PRA & LRA \\
\midrule
FCNCD\_EB&	0.597&	0.277&	0.678&	0.356& 0.771&	0.357\\
FCNCD\_BPR&	0.594&	0.261&	0.672&	0.332& 0.742&	0.307\\
FCNCD\_List&	0.584&	0.272&	0.669&	0.332& 0.726&	0.284\\
FCNCD\_MO&  {0.598}& {0.278}&  {0.679}&  {0.368}&  {0.772}&  \textbf{0.359}\\
FCNCD&\textbf{0.598}&\textbf{0.279}&\textbf{0.679}&\textbf{0.369}& \textbf{0.772}&	{0.357}\\
\bottomrule
\end{tabular}
\end{center}
\vskip -0.1in
\end{table}

\subsection{Explainability of Diagnostic Participant States (RQ4)}
Monotonicity is a fundamental prerequisite for cognitive diagnosis, and the interpretability of a model is dependent on its adherence to this assumption. According to the monotonicity assumption \cite{rosenbaum1984}, a participant's potential trait ability in a given dimension should be positively correlated with their scores on related items. However, forced-choice tests only provide ranked scores between items, so specific response scores are unavailable. As a result, this study defines the interpretability of the forced-choice test model as follows: a participant with a higher potential trait ability in a dimension is more likely to be ranked higher within the item block.

We use consistency as an evaluation metric for sorting performance, defining the Degree of Agreement (DOA) as follows:
\begin{equation}
    \text{DOA} = \frac{1}{\binom{K}{2}} \sum_{1 \leq a < b \leq K} \mathbb{I}_{\text{DOA}} \big( (F_a^{\text{prof}} > F_b^{\text{prof}}) \land (S_a > S_b) \big),
\end{equation}

where $F^{prof}_a$ and $F^{prof}_b$ represent the potential ability values of the participant in dimensions $c_a$ and $c_b$, respectively, and $S_a$ and $S_b$ denote the sums of ranked scores for the items in those dimensions. The condition holds when the potential ability in dimension $c_a$ exceeds that in dimension $c_b$, along with a corresponding greater sum of ranked scores for dimension $c_a$.
\begin{figure}[ht]
\vskip 0.2in
\begin{center}
\centerline{\includegraphics[trim=6cm 5cm 6cm 4.3cm, clip,scale=0.8]{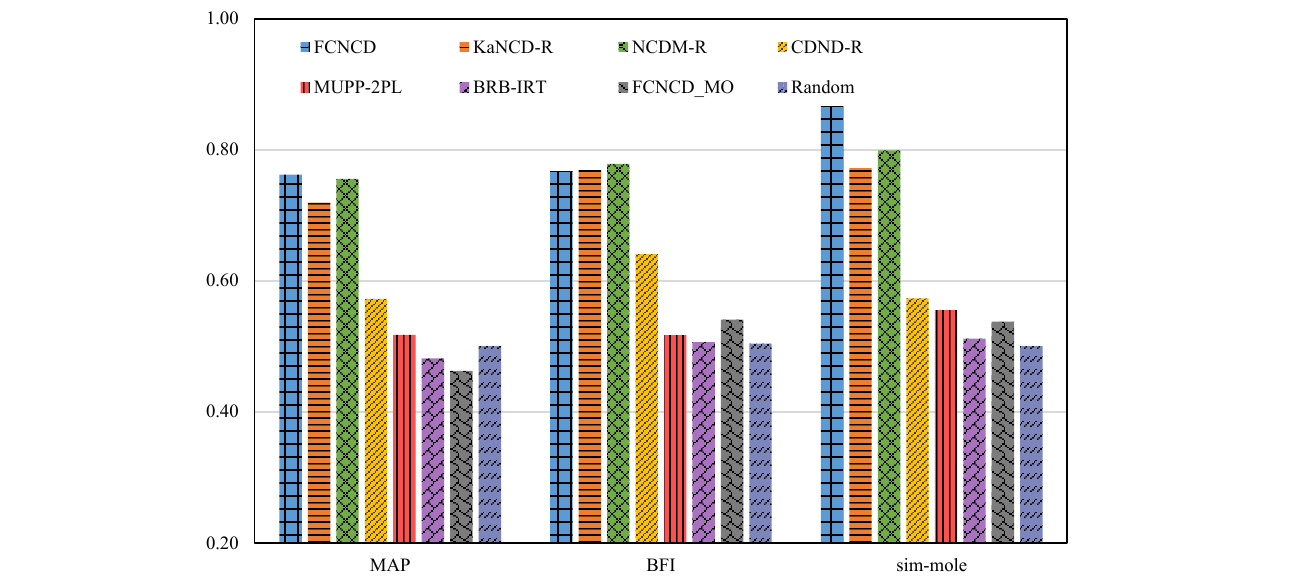}}
\caption{DOA results of models on three datasets.}
\label{Explainability}
\end{center}
\vskip -0.2in
\end{figure}

We evaluate the FCNCD model against several benchmark models, including KaNCD-R, NCDM-R, CDND-R, MUPP-2PL, BRB-IRT, FCNCD\_MO, and Random. Notably, MF and RankNet are excluded from the comparison due to the lack of clear correspondences between their latent characteristics and knowledge concepts. Figure \ref{Explainability} illustrates the experimental results:
(1) FCNCD, KaNCD-R, and NCDM-R consistently rank in the top three for DOA values across all three datasets. FCNCD performs best in the MAP and sim-mole datasets, followed by NCDM-R and KaNCD-R, while their performance in the BFI dataset is similar, with NCDM-R slightly outperforming the others.
(2) Despite being a deep learning model, CDND-R's DOA values are lower than those of other deep learning models while remaining higher than the random baseline. 
(3) The remaining baseline models' DOA values are similar to the random baseline, indicating limited consistency in ranking latent abilities.
(4) The interpretability of FCNCD\_MO, the variant without the monotonicity assumption, approaches the random baseline—a finding consistent with prior studies \cite{WangNCD2020,wang2022neuralcd}.

In conclusion, the FCNCD model not only accurately captures latent ability traits, but it also effectively models the ordinal relationships between items, thereby providing strong support for the interpretability of forced choice tests.

\subsection{Case Study (RQ5)}

\begin{figure*}[ht]
\vskip 0.2in
\begin{center}
\centerline{\includegraphics[trim=5cm 6.6cm 6cm 6cm, clip,scale=0.65]{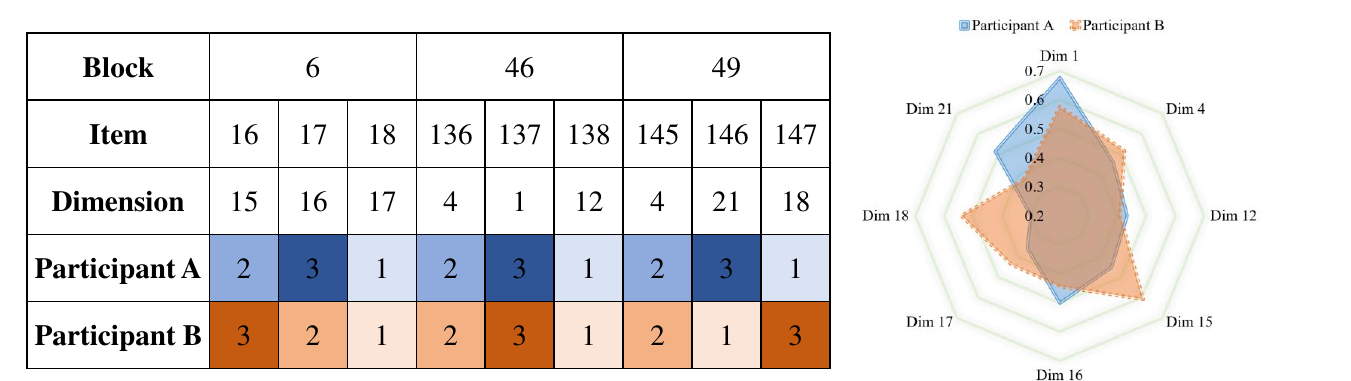}}
\caption{Diagnosed results of two participants on MAP.}
\label{fcncd_casestudy}
\end{center}
\vskip -0.2in
\end{figure*}

Figure \ref{fcncd_casestudy} depicts a specific case of two participants diagnosed by the FCNCD model in the MAP dataset. The left side of the figure depicts the participants' responses across three item blocks, including the corresponding item dimensions and sorted responses. The right side displays radar charts depicting the participants' potential competence values in these dimensions.

Several conclusions can be drawn from the graphs: \textbf{(1)} Participants with higher ability values in a dimension are more likely to rank items in that dimension favorably. For example, in item block 46 (which included items 136, 137, and 138, which correspond to dimensions 4, 1, and 12, respectively), both participants ranked the items as "2, 3, 1", indicating that they thought item 137 (dimension 1) was the most relevant to their situation. \textbf{(2)} Although Participants A and B had the same rankings in item block 46, there are significant differences in their potential ability values for the pertinent dimensions. Participant A has a significantly higher ability value than Participant B in dimension 1, while the opposite is true in dimension 4; in dimension 12, the trends are reversed. This case demonstrates the FCNCD model's ability to estimate potential trait abilities at the micro level, as well as its effectiveness in inferring individual differences in ability traits using response ranking.

\section{Conclusion}
FCNCD is a cognitive diagnostic model that predicts potential traits and response performance in forced-choice tests. The model starts by characterizing both participants and items, then uses the unidimensionality of forced-choice item blocks to extract participants' potential abilities based on the item dimensions. It takes into account both item difficulty and discrimination. To improve feature extraction, FCNCD performs nonlinear mapping on the embedded representations of participants and items. An interaction function captures the combined effects of participant ability, item difficulty, and discrimination, and final predictions are made using a fully connected layer. Recognizing the distinction between forced-choice ranking tasks and recommender system tasks, we modify the model's optimization loss function to better fit the forced-choice context. Extensive experiments using two real-world datasets and one simulated dataset show that the FCNCD model produces accurate and interpretable results for cognitive diagnosis in forced-choice tests.

Future research can extend in a variety of ways: \textbf{(1)} Using participant response times (RT) to investigate their impact on decisions. Response time reflects participants' cognitive processes and correlates with both their cognitive abilities and item difficulty, making it an important input feature for improving diagnostic accuracy. \textbf{(2)} Enhancing item features with item-specific text information. Currently, item features are primarily based on static attributes such as dimensionality and difficulty; investigating the semantic content of item text may improve prediction performance. \textbf{(3)} Using deep learning techniques to improve computerized adaptive test selection algorithms for forced-choice questions. Deep learning can lead to more intelligent item selection strategies, improved test personalization and adaptability, and, ultimately, a more accurate assessment of participants' cognitive abilities.

\bibliographystyle{unsrt}  
\bibliography{references}

\end{document}